\documentclass{article}
\usepackage{multirow}

\usepackage{PRIMEarxiv}

\usepackage[utf8]{inputenc} 
\usepackage[T1]{fontenc}    
\usepackage{hyperref}       
\usepackage{url}            
\usepackage{booktabs}       
\usepackage{amsfonts}       
\usepackage{nicefrac}       
\usepackage{microtype}      
\usepackage{lipsum}
\usepackage{fancyhdr}       
\usepackage{graphicx}       
\graphicspath{{media/}}     

\title{Connective Reconstruction-based Novelty Detection}
\author{
  Seyyed Morteza Hashemi \\
   Institute for Advanced Studies in Basic Sciences (IASBS)\\
  \texttt{MortezaHashemi@iasbs.ac.ir} \\
   \And
  Parvaneh Aliniya \\
  University of Nevada, Reno \\
  \texttt{Aliniya@nevada.unr.edu} \\
  \And
  Parvin Razzaghi \\
   Institute for Advanced Studies in Basic Sciences (IASBS) \\
  \texttt{P.razzaghi@iasbs.ac.ir} \\
}

\begin{document}
\maketitle

\begin{abstract}
Detection of out-of-distribution samples is one of the critical tasks for real-world applications of computer vision. The advancement of deep learning has enabled us to analyze real-world data which contain unexplained samples, accentuating the need to detect out-of-distribution instances more than before. GAN-based approaches have been widely used to address this problem due to their ability to perform distribution fitting; however, they are accompanied by training instability and mode collapse. We propose a simple yet efficient reconstruction-based method that avoids adding complexities to compensate for the limitations of GAN models while outperforming them. Unlike previous reconstruction-based works that only utilize reconstruction error or generated samples, our proposed method simultaneously incorporates both of them in the detection task. Our model, which we call "Connective Novelty Detection" has two subnetworks, an autoencoder, and a binary classifier. The autoencoder learns the representation of the positive class by reconstructing them. Then, the model creates negative and connected positive examples using real and generated samples. Negative instances are generated via manipulating the real data, so their distribution is close to the positive class to achieve a more accurate boundary for the classifier. To boost the robustness of the detection to reconstruction error, connected positive samples are created by combining the real and generated samples. Finally, the binary classifier is trained using connected positive and negative examples. We demonstrate a considerable improvement in novelty detection over state-of-the-art methods on MNIST and Caltech-256 datasets.
\end{abstract}


\section{Introduction}
Novelty detection identifies whether a given sample is out-of-distribution or within the training data distribution. Out-of-distribution instances are referred to as novel instances (negative), which the method does not have access to them during the training. It is closely related to topics such as anomaly detection, outlier detection, and open set recognition.
Thanks to the emergence of the powerful computational system and the ever-progressing deep learning approaches, the research community is recently moving forward to designing approaches to leverage real-world and online datasets rather than predefined ones. This, in turn, has brought novelty detection into the spotlight in the recent decade because one of the main challenges of using real-world datasets is that they contain many uncaptured or poorly represented samples.

As a result, models will not be able to learn the accurate distribution of the training data (positive class). Therefore when presented with new examples, they will fail to determine whether this unexplained example is from the training data distribution.

One common way to evaluate the performance of a novelty detection model is to select a category as a positive class for training a classifier, and in the test stage, use a combination of samples from this class and one or several other classes (negative classes). In this way, the performance of the model demonstrates its ability to distinguish new samples within the positive class from others. Having only access to a positive set in training means that the performance of a model is closely related to its ability to learn the representation of the positive class. One approach is using a one-class classifier that learns the representation of the positive class such as one class SVM~\cite{scholkopf1999support,wang2004anomaly, chen2013new} and GAN-based methods~\cite{zenati2018adversarially, han2021gan,schlegl2017unsupervised}. In the test phase, query samples with low classification scores for the positive class will be labeled as negative. Another interesting direction is leveraging autoencoder to reconstruct the training examples in a GAN-based model~\cite{Sabokrou_2018_CVPR, schlegl2019f, akcay2018ganomaly, zhou2020sparse, sakurada2014anomaly,collin2021improved}. In this methods, the reconstructor tries to generate an authentic version of the input, and the classifier attempts to distinguish between actual and reconstructed samples. A critical property of this variation of a one-class classifier is that as the reconstructor is trained using positive samples, it will destroy negative examples in the test phase, making it easier for the classifier to discriminate between two classes.

Our proposed model is close to the latter approach in terms of using reconstruction to make the negative samples more distinguishable from the positive ones; however, there are several downsides that we addressed in a new formulation of the problem by using a binary classifier in a non-GAN setting. The motivation behind this modification comes from the fact that the significant performance improvement of these approaches comes from destroying the negative samples in the reconstruction phase. In addition, training a GAN model is entangled with specific challenges. In order to formulate the method as a GAN model, the input of a classifier will be reconstructed image, limiting learning the representation of the positive class to the accurateness of generated images. Therefore we model a reconstructor as a denoising autoencoder (Unet~\cite{ronneberger2015u}) by customizing vanilla Unet to the task in a way that this new version improves the reconstruction process for the positive samples and deteriorates it for negative ones. Also, the denoising version makes it robust to noise.
In the case of the traditional one-class classifier, the model only distinguishes the sample of two classes based on their similarity to the learned positive class’ characteristics, as it is only trained on positive samples. However, if the model has access to the negative class, it would develop a more accurate classification boundary and classify a test image by considering the learned properties of both classes. We introduce a procedure to create negative samples via manipulating the positive samples. While forcing them to differentiate from the positive instances, the proposed method keeps negative samples close enough to positive ones in order to create hard negative examples, enabling the model to form a more accurate boundary between the classes. Therefore we leverage a binary classifier that is trained on two positive and generated negative examples.

\begin{figure}
  \centering
  \includegraphics[width = 0.8 \columnwidth]{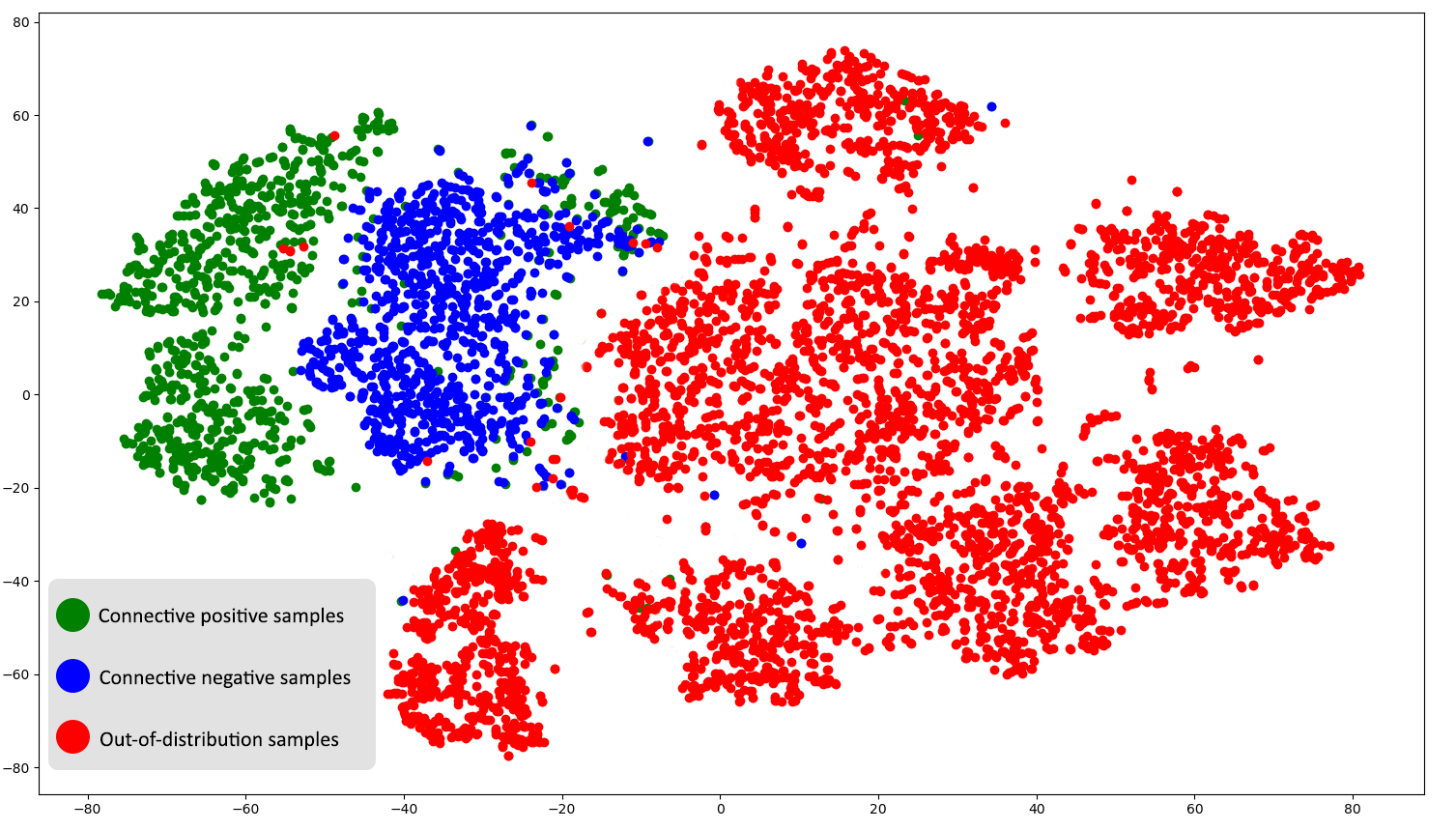}
  \caption{
The t-SNE visualization of the positive, generated hard negative and out-of-distribution samples for the MNIST dataset.
 Distributions of generated hard negative examples are very close to those of positive samples. With proper training of the classifier using positive and generated negative samples, the classifier learns the accurate boundaries to distinguish between positive and out-of-distribution samples. As a result, out-of-distribution samples can be recognized with a higher probability in the evaluation phase.
  }
  \label{tsne}
\end{figure}

Negative samples are created by applying data augmentation techniques to the reconstructed positive samples.To make the negative examples “hard”, the resulting image is combined with the original one by a weighted summation, creating “connective negative examples”. As the reconstructor is trained separately, the reconstruction error for the positive examples might propagate to training the classifier; reconstructed positive samples are also combined with original images, in the same way, forming “connective positive samples” to mitigate error propagation. By creating connective positive and negative examples, we also incorporate reconstruction error in the training of the classifier. For positive classes, as the input and generated image are close, the error will become smaller than using only generated image, so the combination will improve the features, while for the negative examples, the opposite is the case. Figure~\ref{tsne} represents the distribution of the positive samples, generated hard examples, and out-of-distribution samples generated using t-SNE~\cite{van2008visualizing}.

The main contributions of our method are listed below: 
\begin{enumerate}
  \item Proposing a new efficient reconstruction-based approach. 
  \item Customizing the autoencoder network to boost its performance for positive samples and decline it for negative examples as it is favorable for the novelty detection task. 
  \item Introducing a procedure to generate hard negative examples which make the binary classification's boundary more precise. 
  \item Using the idea of connective positive and negative samples generation to train a binary classifier with the goal of alleviating error propagation for positive examples and making negative examples hard. 
  \item Outperforming state-of-the-art methods on MNIST and Caltech 256 datasets.
  
\end{enumerate}

The next sections of the paper are structured as follows. First, Section 2 presents a detailed review of the previous works in anomaly detection that are closely related to this paper. Section 3 covers the motivation and development of the proposed method. The experimental setup, evaluation metric, and results and analysis are described in Section 4. Finally, we conclude in Section 5 with a summary and future directions for future work.

\section{Related Works}
Novelty detection is closely related to outlier and anomaly detection. For anomaly detection, out-of-distribution data can be used in the training process. In contrast, the training process of novelty detection is unsupervised as the out-of-distribution data is not presented in the training phase. One primary approach is to learn the latent space representation of the target class and then reject dissimilar instances ~\cite{you2017provable,xia2015learning,chalapathy2018anomaly,ruff2018deep}. These approaches are one-class classifications for which it is difficult to form an effective feature representation of the positive class, especially for high-dimensional data, such as images.

Reconstruction-based approaches are another research direction, addressing novelty detection by assuming that training set and novel examples are from standard classes; therefore, reconstruction error can be used as a signal for detecting out-of-distribution examples or training a classifier ~\cite{sakurada2014anomaly, huang2019inverse, collin2021improved}. In the former category, Mayu et al. in~\cite{sakurada2014anomaly} model anomaly detection using an autoencoder as a dimensionality reduction technique. Assuming that the inlier class has correlated data features when embedded in low dimension space, this correlation makes the inlier and outlier significantly separable; therefore, reconstruction error is directly used as an abnormality score in the test phase.  In~\cite{huang2019inverse}, the inverse-transform autoencoder is first trained using a set in which specific target information is removed based on a set of transformations to embed corresponding erased information during the reconstruction and learn a better representation. Collin et al.~\cite{collin2021improved} used the synthetic noise (randomly colored ellipse with irregular edges) to corrupt the input to prevent the autoencoder with skip connections (AESc) from learning the identity map.

The second category that uses the reconstructed samples in training consists of primarily GAN-based models. Proposed method in~\cite{Sabokrou_2018_CVPR} provides an end-to-end architecture for identifying novelty classes. An adversarial model architecture typically consists of two subnets, one generator and one discriminator. An autoencoder model is used in this architecture as the generator. The generator attempts to reconstruct positive examples during the training phase by learning the distribution of the target class. Consequently, it is not able to reconstruct negative samples. Using both subnets in the evaluation phase allows the generator to help the discriminator identify negative samples by destroying them. 

The WGAN~\cite{arjovsky2017wasserstein} approach learns to produce samples in the desired distribution during the training phase. Since no negative samples are available, Pourreza and his colleagues attempted to create negative samples using WGAN in G2D~\cite{Pourreza_2021_WACV}.The main idea of this work is to use samples generated in the first epochs that generally do not resemble the training samples as irregularity. By measuring the generated samples at the end of each epoch, G2D determines a threshold T and considers the samples generated from the first epoch to the $T$ epoch as negative samples. $T$ is the threshold between negative and positive samples generated by WGAN. In order to detect the novelty samples, they trained a binary classifier with the positive and the negative generated samples. 

OLED~\cite{jewell2022one} learns an effective semantically reach representation by generating optimal masks for context autoencoder using adversarial training. In this setting, the mask module tries to generate optimal masks while the reconstructor module attempts to generate original images from masked input. Learning optimal masks enhances the novelty detection results. In SCADN~\cite{yan2021learning}, YAN et al. create a reach semantic contextual cues by removing parts of the inputs using the multi-scale stripped binary mask, then a generative adversarial network is utilized to reconstruct the unseen regions. 

Our work is similar to both types of reconstruction-based approaches from specific perspectives. While having two parts for reconstruction and discrimination, it is not trained in an adversarial manner but defined to utilize reconstructed images in the classifier’s training. Also, it resamples the first group as it uses the reconstruction error in creating connective instances. We introduce new concepts in a unified framework to benefit from their advantages and surpass the limitations of separate views.

\section{Proposed Method}
\begin{figure}
  \centering
  \includegraphics[width = 1 \columnwidth]{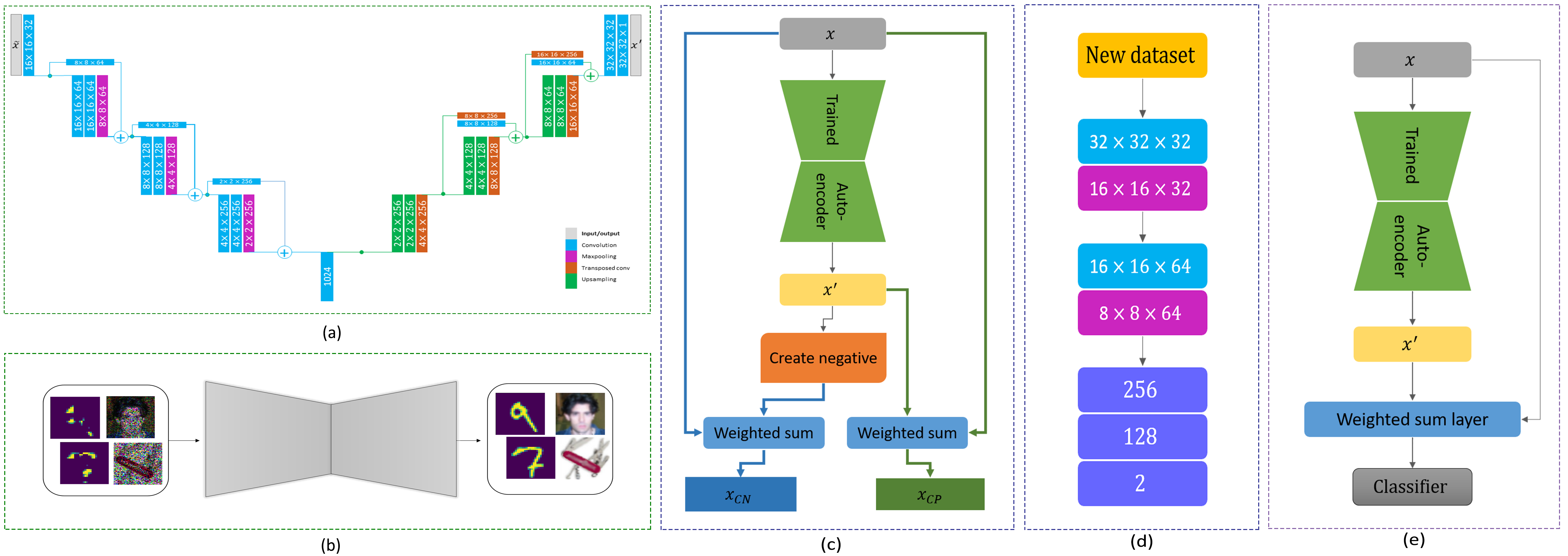}
  \caption{Proposed method architecture and how to train it. (a) Refined Unet as the reconstruction sub-network in the proposed architecture. (b) The autoencoder is trained using augmented samples and learns the positive class distribution by reconstructing this samples. (c) The connective positive samples are created by applying a weighted sum of the input positive samples and their corresponding reconstructed samples. The connective negative samples are constructed by applying weighted sum of the positive samples and the corresponding augmented reconstructed samples. (d) The convolutional classifier is trained using the newly generated dataset. (e) After the training process, the weights of the autoencoder and classifier models are replaced in the proposed architecture.}
  \label{models}
\end{figure}

Our proposed novelty detection method (Figure~\ref{models}-e) has two modules that are trained separately. The first module, reconstructor, is an autoencoder that learns the representation of the positive class via learning the latent space of this class. New connected positive and negative samples will be generated in the second phase using the reconstructor’s outputs. Finally, a binary classifier, a convolutional neural network, will be trained using this new dataset. Figure~\ref{models} summarizes the architecture of the method and the training procedure. In the following, each module of  Figure~\ref{models} is explained in detail.

In Figure~\ref{models}-a, the architecture for the autoencoder is presented. The overall design of the reconstructor is similar to the UNet, with one modification that makes it suitable for our task. Instead of using a long-skip connection from the encoder to the decoder, we used local short-skip connections by forming residual blocks in the encoder and decoder. The main reason for this stem from the fact that in novelty detection based on a reconstructor, while the generated image for positive samples should resemble the original images, for out-of-the-distribution samples, it should destroy them to have high reconstruction error. The benefit of using the long-skip connection in the Unet is to use fine-grained, low-level features of the encoding phase when reconstructing them in the decoder. Therefore it will promote the performance of the reconstructor for both positive and negative samples in the test phase, which is unfavorable for the task at hand. So we replaced the long-skip connection with a short-skip connection which will help with faster convergence. The coarse-dropout augmentation technique and gaussian noise are applied to the input of the reconstructor to make the model better learn the distribution of positive samples and how to reconstruct them (Figure~\ref{models}-b). It is worth mentioning that avoiding long-skip connections prevents noise propagation to the reconstruction of the input. The reconstructor is trained on the positive class to learn the representation of this class.

The second phase is generating connected positive and negative samples. Traditionally, reconstruction models either train the model on positive examples and, in the test phase, use the reconstruction error to determine if the query sample is out-of-distribution or use the generated sample in training a one-class classifier. The shortcoming of the first category is that if the autoencoder is trained well, the difference between the error for positive and out-of-distribution samples will be small, adversely impacting the detection. The downside of the latter category lies in the fact that they are primarily GAN-based models, so they should alleviate the training instability and mode-collapse issues in GANs by limiting the design to specific models.

In this work, we address these problems by generating a new dataset consisting of connective positive and negative samples only from the positive class and training a binary classifier using these data. In contrast to one class classifier that only relays on the given positive samples to discriminate them from unexplained images in the test time, our model predicts the classification boundary more precisely by creating negative samples generated to have close distribution to the positive samples.
The process of creating connective positive and negative samples is shown in Figure~\ref{models}-c. Connective positive samples are created to combine the raw sample with the reconstructed one with a weighting hyperparameter; as the training of the classifier is separated from the reconstructor, this will make the classifier robust to the reconstruction error (Figure~\ref{models}-d). 
For the connective negative examples, one of the two operations, motion blur and Gaussian noise is applied to one of the original or reconstructed samples so that it would be considered from an unseen distribution. The raw sample is combined with the resulting image to maintain the closeness of negative examples. This step can be considered as creating hard negative examples that would endow the method with the ability to reject out-of-distribution examples close to the positive samples, resulting in better accuracy for query images close to the positive class that are truly from a different distribution. In the final step, the binary classifier will be trained with connected positive and negative samples labeled as members of two classes. 

In the test stage, for a query image, first, a reconstructed version will be created, then the query image and the regenerated image will be combined and considered as the input for the classifier (Figure~\ref{models}-e). In the next section, the details of the training is presented. 

\subsection{Training Phase}
For every image in the training set $S=[x]_1^M$, consisting of $M$ samples, Gaussian noise will be added to it in order to create the input of the autoencoder; then, coarse dropout will be applied to it. Finally, $x_{c}$ will be used to train the autoencoder.

\begin{equation}
	x_{g}=x+\eta, \eta\sim N(0,\mu)
\end{equation}

\begin{equation}
    x_{c}=x_{g}-\beta \odot x_{g}
\end{equation}
The main idea of the coarse-dropout augmentation is to randomly remove rectangles of pixels from image’s channels. $\beta$ is a matrix with same number of channels and size of $x_{g}$ with pixel values of one for pixels within the rectangles and zero otherwise. The number of rectangles and their sizes are determined by two hyperparameters of dropout probability and low-resolution image size. 

The training loss for the autoencoder is defined as follows:
\begin{equation}
    L=\sqrt{(x-x^\prime)^2+\varepsilon^2}
\end{equation}\label{charloss}

It is Charbonnier loss that behaves like  $L2$ near the origin and like $L1$ far from the origin.
In the formula~\ref{charloss} $x^\prime$ is the output of the autoencoder. In the loss function, rather than calculating the difference between the input (original image) and reconstructed version, it compares the original image with the output, enhancing the robustness of the model to noise in the positive class. After training the autoencoder, as shown in Figure~\ref{models}-c connected positive and negatives examples will be generated by:

\begin{equation}
    x_{CP}=\lambda x+(1-\lambda)x^\prime
\end{equation}

\begin{equation}
    x_{CN}=\lambda x+(1-\lambda)x_d^\prime
\end{equation}

in which $x_{CP}$ and $x_{CN}$ are connected positive and negatives examples respectively. $\lambda $ is a hyperparameter to determine the importance of each of the original and reconstructed samples.
$x_d^\prime $ is the generated negative example obtained by applying motion blur or Gaussian noise to the input image.
For each sample corresponding class label (positive or negative) will be considered as the target value to be used in the classifier.  Therefore connective positive and negative set will become $S_P=[(x_{CP},1)]_{1}^{M}$ and $S_N=[(x_{NP},1)]_{1}^{M}$. In this way, the training set will be $S_A={S_{CP},S_{CN}}$. The classifier in Figure \ref{models}-d will be trained with this new dataset.

\subsection{Test Phase}
In the test stage, every query image will be processed, as shown in  Figure \ref{models}-e. First, the autoencoder will regenerate the new version based on the learned latent representation. Then, connective samples will be formed as the input of the classifier.

\begin{equation}
    f(x) = C (\lambda x + (1 - \lambda )A(x))
\end{equation}

Where $A$ refers to the autoencoder and $C$ refers to the classifier.

\section{Experimental Results}
This section presents the experimental results of evaluating the proposed method on two commonly used datasets for novelty detection. We compare our results with state-of-the-art approaches. Standard metrics in novelty detection, AUC-ROC, and F1 score are used to evaluate the models’ performance. A batch size of 128 is used, and the weighting coefficient for connective example creation is set to 0.5 experimentally.

\subsection{Results for MNIST dataset}
\begin{figure}
  \centering
  \includegraphics[width = 0.5 \columnwidth]{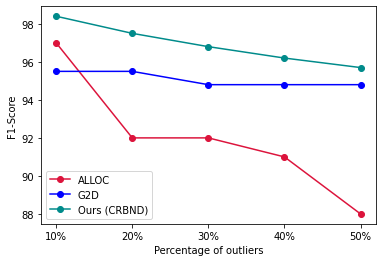}
  \caption{Comparison of F1 score for the proposed method, ALLOC, and G2D for different outlier percentages for MNIST dataset.}
  \label{mnistF1}
\end{figure}

This dataset \cite{deng2012mnist} consists of 60,000 training and 10,000 test images that are 28 by 28 pixel grayscale images of handwritten digits. The model is trained using each of the ten classes as the positive class once; random sampling of examples of the same and other classes is considered for testing. We tested with 10-50 percent of negative classes; the average results for all ten classes are reported in diagram~\ref{mnistF1}. For comparison, we used ALLOC and G2D. ALLOC is a GAN-based one-class classification approach, and G2D is also a GAN-based model, which is a binary classifier. As shown in diagram~\ref{mnistF1}, our model’s performance is superior with a large margin in all cases compared to the state-of-the-art novelty detection models. It should be noted that there is a consistency in the performance decrease with negative class percentage increment.

\subsection{Results for Caltech 256 dataset}
This dataset \cite{griffin2007caltech} contains 30,607 images from 256 classes. Each category has at least 80 images. We repeat the testing procedure 20 times and use images from n $\in$ {1, 3, 5} randomly chosen categories as the positive class. It should be noted that a combination of three and five classes as the positive class could be considered as multi-class novelty detection. We select the “clutter” category as the negative class in testing, such that each experiment has exactly 50\% outliers (the number of examples in classes is different, so it varies accordingly). Table \ref{caltech-256_tableRes} summarizes the results for the proposed method and two state-of-the-art approaches. Our model outperforms both one-class and binary classifiers. The improvement in the F1-score and AUC-ROC for all three variations of the positive class is consistent and superior to previous works.

\begin{table}
\begin{center}
\begin{tabular}{|l|ll|ll|ll|}
\hline
\multirow{2}{*}{Method} & \multicolumn{2}{c|}{One class} & \multicolumn{2}{c|}{Three classes} & \multicolumn{2}{c|}{Five classes} \\
 & \multicolumn{1}{c}{AUC} & \multicolumn{1}{c|}{F1} & \multicolumn{1}{c}{AUC} & \multicolumn{1}{c|}{F1} & \multicolumn{1}{c}{AUC} & \multicolumn{1}{c|}{F1} \\ \hline\hline
ALLOC~\cite{Sabokrou_2018_CVPR} & \multicolumn{1}{l|}{94.2} & \multicolumn{1}{l|}{92.8} & \multicolumn{1}{l|}{93.8} & \multicolumn{1}{l|}{91.3} & \multicolumn{1}{l|}{89.5} & 90.5 \\
G2D~\cite{Pourreza_2021_WACV} & \multicolumn{1}{l|}{95.7} & \multicolumn{1}{l|}{94.5} & \multicolumn{1}{l|}{95.1} & \multicolumn{1}{l|}{90.0} & \multicolumn{1}{l|}{93.9} & 91.3 \\

R-graph~\cite{you2017provable} & \multicolumn{1}{l|}{94.8} & \multicolumn{1}{l|}{91.4} & \multicolumn{1}{l|}{92.0} & \multicolumn{1}{l|}{88.0} & \multicolumn{1}{l|}{91.3} & 85.8 \\

REAPER~\cite{lerman2015robust} & \multicolumn{1}{l|}{81.6} & \multicolumn{1}{l|}{80.8} & \multicolumn{1}{l|}{79.6} & \multicolumn{1}{l|}{78.4} & \multicolumn{1}{l|}{65.7} & 71.6 \\

OutlierPursuit~\cite{xu2010robust} & \multicolumn{1}{l|}{83.7} & \multicolumn{1}{l|}{82.3} & \multicolumn{1}{l|}{78.8} & \multicolumn{1}{l|}{77.9} & \multicolumn{1}{l|}{62.9} & 71.1 \\

LRR~\cite{liu2010robust} & \multicolumn{1}{l|}{90.7} & \multicolumn{1}{l|}{89.3} & \multicolumn{1}{l|}{47.9} & \multicolumn{1}{l|}{67.1} & \multicolumn{1}{l|}{33.7} & 66.7 \\

CRBND (Ours) & \multicolumn{1}{l|}{\textbf{99.8}} & \multicolumn{1}{l|}{\textbf{98.4}} & \multicolumn{1}{l|}{\textbf{99.6}} & \multicolumn{1}{l|}{\textbf{98.0}} & \multicolumn{1}{l|}{\textbf{99.4}} & \textbf{97.8} \\ \hline
\end{tabular}
\end{center}
\caption{Comparison of F1 score and AUC-ROC for the proposed method and top researches in the field. The first set of F1-score and AUC-ROC columns, corresponds to the performance of the methods when one class was considered as positive in training, and the next ones are for three and five classes as the positive class.}
\label{caltech-256_tableRes}
\end{table}

\begin{figure}
  \centering
  \includegraphics[width = 0.7 \columnwidth]{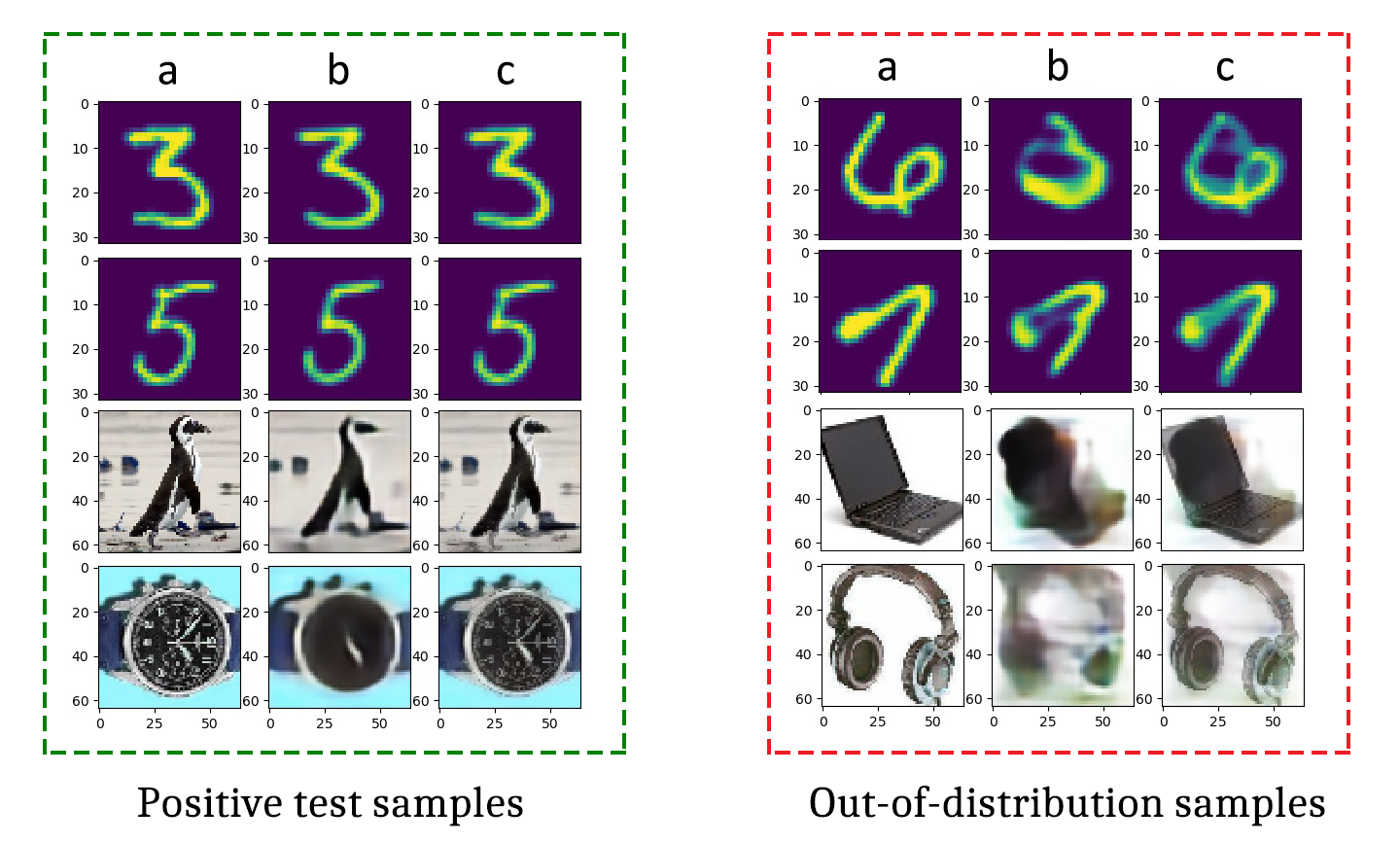}
  \caption{Examples of original (a), reconstructed (b) and connective (c) images of the proposed method for MNIST and Caltech-256 datasets.}
  \label{samples_mnist}
\end{figure}

We contribute these improvements, which are valid for both datasets, to our new formulation of the problem, which eliminates the limitations of GAN-based models, and to the fact that the procedure of generating hard negative samples with close proximity to the positive class in our work endows the model with considerable robustness compare to the state-of-the-art approaches, ALLOC and G2D. Examples in Figure \ref{samples_mnist} for MNIST and Caltech-256  datasets show the performance of the reconstructor module and connective approach. The reconstruction error is far less for positive test images compared to out-of-distribution examples. For instance, the boundaries of numbers and objects are more well-defined in the positive test images in Figure \ref{samples_mnist}-b; however, for the out-of-distribution instances, they are destroyed. One key observation is that the power of the reconstructor is to change the out-of-distribution samples so that they no longer belong to the normal distribution. Results of connective examples in the last column of Figure \ref{samples_mnist}-c  show that for positive test images, the method compensates for lost details in the reconstruction phase. For the out-of-distribution examples, the connective operation makes the destroyed image
closer to the original image. They are far enough to be out-of-distribution and close enough to make the training robust as hard negative examples.

\section{Conclusion}
In this paper, we presented a connective reconstruction-based novelty detection approach to enhance the learning of the latent representation of the positive class via connective example creation. In addition, using hard negative examples, we formulated a binary classifier to promote the discriminatory power of the model by encouraging a more accurate decision boundary. The experimental results show the state-of-the-art performance of our model. In future research, we will utilize a more powerful autoencoder by considering contextual information in learning the latent representation of the positive class. For the negative sample generation, we will examine the validity of the idea of using online adversarial training.

\bibliographystyle{unsrt}  
\bibliography{references}

\end{document}